\documentclass[letterpaper, 10 pt, conference]{ieeeconf}  

\IEEEoverridecommandlockouts                             

\usepackage{multicol}
\usepackage[bookmarks=true]{hyperref}
\usepackage{bm}
\usepackage{amssymb}
\usepackage{threeparttable}
\usepackage{booktabs}
\usepackage{wrapfig}
\usepackage{amsmath}
\usepackage{wrapfig}
\usepackage{lipsum}
\usepackage{siunitx}
\usepackage{graphicx}
\usepackage{url}
\usepackage{cleveref}
\usepackage{cite}
\usepackage{times}
\usepackage[table,xcdraw]{xcolor}
\usepackage{multirow}
\usepackage{color}

\begin{document}

\title{\LARGE \bf
Enabling Stateful Behaviors for Diffusion-based Policy Learning}

\author{Xiao Liu, Fabian Weigend, Yifan Zhou, and Heni Ben Amor 
\thanks{All authors are with the School of Computing and Augmented Intelligence, Arizona State University, USA {\tt\small \{xliu330, fweigend,, yzhou298,  hbenamor\}@asu.edu}}%
}



%

\maketitle

\begin{abstract}
While imitation learning provides a simple and effective framework for policy learning, acquiring consistent actions during robot execution remains a challenging task. Existing approaches primarily focus on either modifying the action representation at data curation stage or altering the model itself, both of which do not fully address the scalability of consistent action generation. To overcome this limitation, we introduce the Diff-Control policy, which utilizes a diffusion-based model to learn the action representation from a state-space modeling viewpoint. We demonstrate that we can reduce diffusion-based policies' uncertainty by making it stateful through a Bayesian formulation facilitated by ControlNet, leading to improved robustness and success rates. Our experimental results demonstrate the significance of incorporating action statefulness in policy learning, where Diff-Control shows improved performance across various tasks. Specifically, Diff-Control achieves an average success rate of 72\% and 84\% on stateful and dynamic tasks, respectively. Project page: \url{https://github.com/ir-lab/Diff-Control}
\end{abstract}

\IEEEpeerreviewmaketitle

\section{Introduction}
\label{sec:intro}
Previous studies have explored various approaches to learning behavioral cloning policies, such as directly outputting actions via regression models~\cite{jang2022bc} or utilizing implicit policies~\cite{florence2022implicit}. Notably, diffusion-based policies~\cite{chi2023diffusion} have emerged as a standout choice due to their ability to model multimodal action distributions effectively, leading to enhanced performance.

In practice however, the concern over inconsistency in action representation remains a persistent challenge. Such inconsistencies can lead to noticeable disparities between the distribution of robot trajectories and the underlying environment, thereby limiting the efficacy of control policies~\cite{qian2023robot}. The primary causes of this inconsistency typically stem from the context-rich nature of human demonstrations~\cite{mandlekar2021matters}, distribution shift problems~\cite{ross2011reduction}, and the volatile nature of high-dynamic environments. Previous approaches, such as action chunking~\cite{zhao2023learning} and predicting closed-loop action sequences~\cite{chi2023diffusion}, have been proposed to address this issue. Additionally, Hydra~\cite{belkhale2023hydra} and Waypoint-based manipulation~\cite{shi2023waypoint} modify action representations to ensure consistency. However, these approaches address the problem by altering the action representation without using the actions as is.

Instead, can we learn to explicitly impose temporal consistency by incorporating temporal transitions within diffusion policies? 
In the realm of deep state-space models (DSSMs)~\cite{NEURIPS2018_5cf68969,klushyn2021latent,kloss2021train}, the effective learning of a state transition model enables the identification of underlying dynamic patterns. In this paper, we argue that such deep transition models can easily be integrated into diffusion policies in order to capture temporal action dynamics as a state-space model. This integration makes the policy stateful, thereby increasing robustness and success rates. 

\begin{figure} [t]
    \centering
    \includegraphics[width=\linewidth]{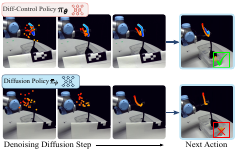}
    \caption{\textbf{Diff-Control Policy} incorporates ControlNet, functioning as a transition model that captures temporal transitions within the action space to ensure action consistency. }
    \label{fig:new_teaser}
\vspace{-0.15in}
\end{figure}


We propose \textbf{Diff-Control}, a stateful diffusion-based policy that generates actions and enables learning an action transition model concurrently. Building upon the ControlNet framework introduced by~\cite{zhang2023adding} for spatial conditioning control in image generation, we leverage it as the transition model to provide temporal conditioning to a base diffusion policy. As shown in \Cref{fig:new_teaser}, a prior action sequence (in blue) is utilized as condition when generating new action sequence (in red). The main contributions of the paper are: 
\begin{itemize}
\item A deep, recursive Bayesian filter within diffusion-based polices using ControlNet structure as a transition model to ensure consistent action generation.
\item Diff-Control stateful policy representation performing dynamic and temporal sensitive tasks with at least 8\% and 48\%  improvement in success rate. 
\item Diff-Control policy exhibits notable precision and robustness against perturbations, achieving at least a 30\% higher success rate compared to state-of-the-art methods.
\end{itemize}

\section{Method}
\label{sec:method}
\begin{figure} [t]
    \centering
    \includegraphics[width=1\linewidth]{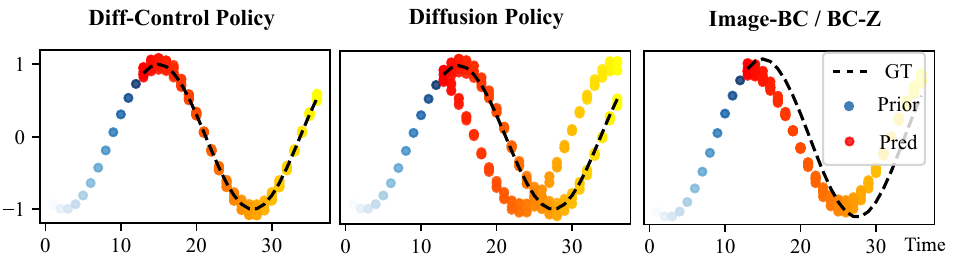}
    \caption{\textbf{Stateful behavior}: at a given state, Diff-Control policy can utilize prior trajectories to approximate the desired function. Diffusion policy~\cite{chi2023diffusion} learns both modes but fails on generating the correct trajectory cosistently, Image-BC/BC-Z~\cite{jang2022bc} fails to generate the correct trajectory.}
    \label{fig:sine}
\vspace{-0.15in}
\end{figure}

The key objective of Diff-Control is to learn how to incorporate state information into the decision-making process of diffusion policies. An illustrative example for this behavior is shown in \Cref{fig:sine}: a policy learning to approximate a cosine function. Given single observation at time $t$, stateless policies encounter difficulties in producing accurate generating the continuation of trajectories. Due to ambiguities, Diffusion Policy~\cite{chi2023diffusion} tends to learn multiple modes. By contrast, Diff-Control integrates temporal conditioning allowing it to generate trajectories by considering past states. To this end, the proposed approach leverages recent ControlNet architectures to ensure temporal consistency in robot action generation.

In computer vision, ControlNet is used within stable diffusion models to enable additional control inputs or extra conditions when generating images or video sequences. 
Our method extends the basic principle of ControlNet from image generation to action generation, and use it as a state-space model in which the internal state of the system affects the output of the policy in conjunction with observations (camera input) and human language instructions.

\subsection{Recursive Bayesian Formulation}
\label{sec:Bayesian_Filter}
The objective of our method is to learn a policy with conditions $\bf c$ and observations $\bf o$ as input. In this context, we define ${\bf a}$ as the trajectory comprising the robot's end-effector pose. In alignment with prior approaches~\cite{chi2023diffusion,zhou2022modularity}, our aim is also to take multiple conditions as input. However, as mention in \Cref{sec:intro}, efforts have been made to explore robust action generation in previous works~\cite{chi2023diffusion,zhao2023learning,belkhale2023hydra}, they have not accounted for the statefulness of ${\bf a}$. We address the action consistency from a Bayesian perspective by introducing transition in action spaces, our formulation is as follows:
\begin{equation}
\begin{aligned}
\label{eq:3}
    p&({\bf a}_t |  {\bf a}_{1:t-1},{\bf o}_{1:t}, {\bf c})\\
    &\propto
    p({\bf o}_t |  {\bf a}_t, {\bf c})\ p({\bf a}_t | {\bf a}_{1:t-1},{\bf o}_{1:t-1}, {\bf c}).
\end{aligned}
\end{equation}
Let $\text{bel}({\bf a}_t) = p({\bf a}_t | {\bf a}_{1:t-1},{\bf o}, {\bf c})$, applying the Markov property, i.e., the assumption that the next generated trajectory is dependent only upon the current trajectory, yields:
\begin{equation}
\begin{aligned}
\label{eq:4}
    \text{bel}({\bf a}_t) = \eta\  \underbrace{p({\bf o}_{t} | {\bf a}_t, {\bf c})}_{\text{observation model}}
    \prod_{t=1}^t \overbrace{ p({\bf a}_t|{\bf a}_{t-1}, {\bf c})}^{\text{ transition model}} \text{bel}({\bf a}_{t-1}),
\end{aligned}
\end{equation}
where $\eta$ is a normalization factor, $p({\bf o}_{t} | {\bf a}_t, {\bf c})$ is the observation model and $ p({\bf a}_t|{\bf a}_{t-1}, {\bf c})$ is the transition model. 

\subsection{Diff-Control Policy}
\label{sec:diff-control_policy}
We now show how Bayesian formulation and diffusion model can be coupled together such that one policy can generate stateful action sequences that facilitate consistent robot behaviors. We propose Diff-Control Policy $\pi_{\pmb{\theta}}({\bf a}_{[W_t]}|{\bf o},{\bf a}_{[W_{t-h}]}, {\bf c})$, which is parameterized by $\pmb{\theta}$. Here, $h$ stands for the execution horizon, ${\bf c}$ represents a language condition in the form of a natural human instruction, and ${\bf o}$ denotes a sequence of images captured by an RGB camera of the scene. The policy $\pi_{\pmb{\theta}}$ generates a window of trajectory ${\bf a}_{[W_t]} = [{\bf a}_1, {\bf a}_2, \cdots {\bf a}_W]^T \in \mathbb{R}^{7\times W}$, where $W$ refers to the window size or the prediction horizon.

\begin{figure}[t!]
\centering
\includegraphics[width=0.85\linewidth]{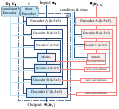}
\caption{The Diff-Control Policy is implemented through the utilization of a locked U-net diffusion policy architecture. It replicates the encoder and middle blocks and incorporates zero convolution layers.}
\label{fig:model}
\vspace{-0.2in}
\end{figure}

The Diff-Control policy within the Bayesian formulation comprises two crucial modules. The transition module receives the previous action ${\bf a}_{[W_t]}$ and generates latent embeddings for the subsequent utilization by the base policy. Acting as the observation model, the base policy incorporates the temporal information associated with ${\bf a}_{[W_t]}$ and produces a new action ${\bf a}_{[W_{t+h}]}$. This two-module structure enables the Diff-Control policy to adeptly capture temporal dynamics and facilitate the generation of subsequent actions with accuracy and consistency.

\textbf{Base Policy}: To begin, we train a diffusion-based policy~\cite{chi2023diffusion} as the base policy $\bar{\pi}_{\pmb{\psi}}({\bf a}_{[W_t]}|{\bf o}, {\bf c})$. We adopt the 1D temporal convolutional networks from~\cite{janner2022diffuser} and construct the U-net backbone. The policy $\bar{\pi}_{\pmb{\psi}}$ can autonomously execute and generate actions without any temporal information dependency.

\textbf{Transition Model}: 
The proposed framework incorporates ControlNet as the Transition Module (depicted in \Cref{fig:model}). This utilization extends the capability of the policy network to include temporal conditioning effectively.
 To achieve this, we utilize the previously generated action sequences as the prompt input to ControlNet. By doing so, the base policy $\bar{\pi}_{\pmb{\psi}}$ becomes informed about the previous actions ${\bf a}_{[W_{t-h}]}$. We implement ControlNet by creating a trainable replica of the $\bar{\pi}_{\pmb{\psi}}$ encoders and then freeze the base policy $\bar{\pi}_{\pmb{\psi}}$. The trainable replica is connected to the fixed model with zero convolutional layers~\cite{zhao2023uni}. ControlNet can then take ${\bf a}_{[W_{t-h}]}$ as the conditioning vector and reuses the trained base policy $\bar{\pi}_{\pmb{\psi}}$ to construct the next action sequence ${\bf a}_{[W_t]}$. 


\subsection{Training}
\label{sec:training}
The training process for the base policy $\bar{\pi}_{\pmb{\psi}}({\bf a}_{[W_t]}|{\bf o}, {\bf c})$ follows a straightforward approach. We firstly encode the observation ${\bf o}$ and language condition ${\bf c}$ into the same embedding dimension, and then we utilize the learning objective defined in \Cref{eq:1} to train the base policy.
    \begin{equation}
    \begin{aligned}\label{eq:1}
    \mathcal{L}_{\text{DDPM}} := \mathbb{E}_{\bf o,a,\tau,z}\left[ \|\epsilon ({\bf o},{\bf a},\tau)-{\bf z}\|_2^2 \right],
    \end{aligned}
   \end{equation}
 The same learning objective is used in finetuning the ControlNet:
    \begin{equation}
    \begin{aligned}\label{eq:5}
    \mathcal{L} := \mathbb{E}_{{\bf o},{\bf c},{\bf a},{\bf a}_{[W_{t}]},\tau,z}\left[ \|\epsilon_{\bf \theta} ({\bf o}, {\bf c},{\bf a}_T,{\bf a}_{[W_{t}]}, \tau)-{\bf z}\|_2^2 \right],
    \end{aligned}
   \end{equation}
where $\mathcal{L}$ is the overall learning objective of the entire diffusion model, $\epsilon_{\pmb{\theta}}(\cdot)$ is the corresponding neural network parameterized by $\pmb{\theta}$. The base policy and the Diff-Control policy are trained end-to-end.

\section{Real-Robot Tasks}
\label{sec:tasks}
We conducted a comprehensive evaluation of the ControlNet Policy by comparing it with four baseline methods across two robot tasks. \Cref{tab:data_summary} provides a summary of the task properties. The tasks encompassed in our evaluation include:
(a) \textbf{Duck Scooping} task in a dynamic scenario, (b) \textbf{Drum Beats} as a periodic task.

\begin{figure}[t]
    \centering
    \includegraphics[width=.95\linewidth]{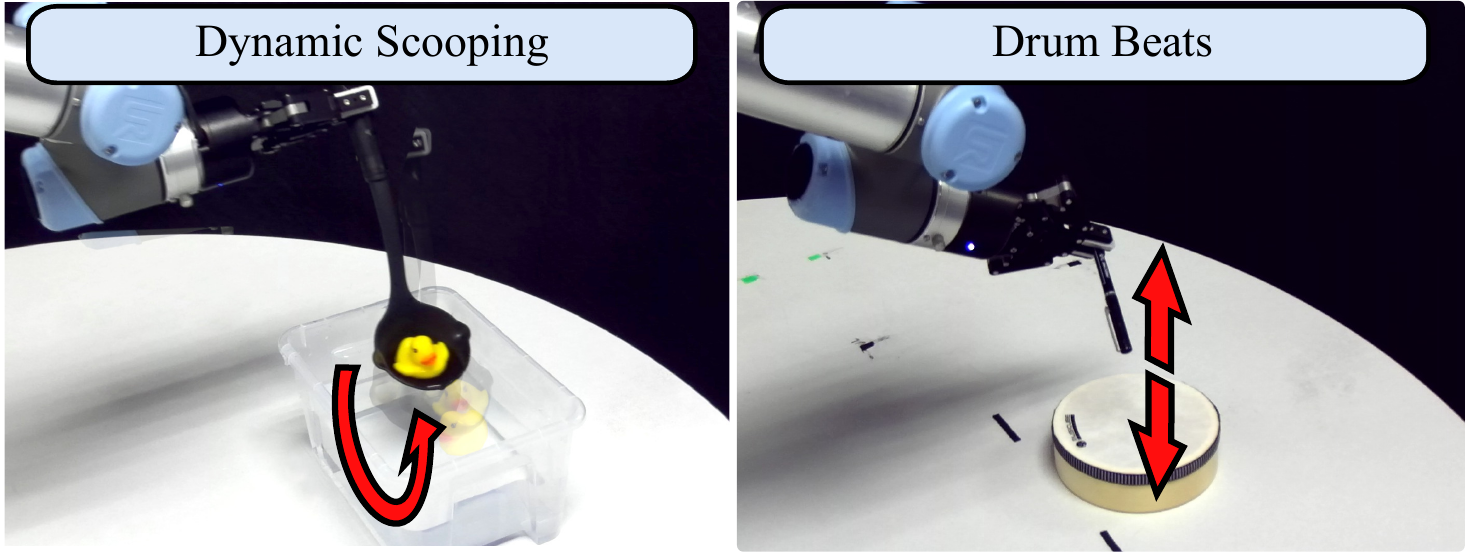}
    \caption{Real-world tasks in this study: a) ``Duck Scooping" task in a water tank, b) ``Drum Beats" task by hitting the drum 3 times.}
    \label{fig:task_teaser}
\end{figure}

\begin{table}[t]
    \begin{center}
    \footnotesize
    \caption{Task Properties.}
    \scalebox{1}{
    \label{tab:data_summary}
    \begin{tabular}{l r c S[table-format=3.0] c r }
    \toprule
        Task  & Dis.  & HiPrec &  \multicolumn{1}{c}{Dem.} & Act. & Steps \\
    \midrule

        Duck Scoop & 0 & $\checkmark$ & 50 & 1 & $\sim$70 \\
        Drum Beats$\times$3 & 0  & $\times$ & 150 & 1 & $\sim$70\\
    \bottomrule
    \end{tabular}}
    \end{center}
\vspace{-0.15in}
\end{table}

\begin{figure*}[t]
\centering
\includegraphics[width=\linewidth]{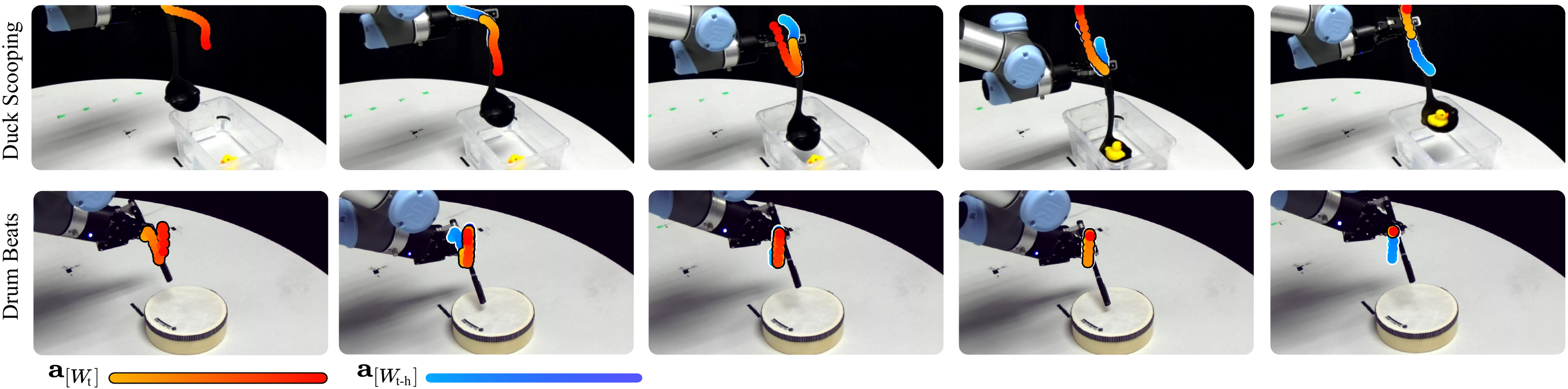}
\caption{\textbf{Diff-Control for real-world tasks}: The first row shows a successful duck scooping experiment. The second row displays one drum task result. The results are best appreciated with videos on the website: \url{https://diff-control.github.io/}.
}
\label{fig:task_plot}
\vspace{-0.1in}
\end{figure*}

The action of the UR5 robot arm is represented as ${\bf a}_{[W_t]}$, where each action is denoted as \mbox{${\bf a}_{i} = [x,y,z,r,p,y,g]^T$}, where $i \in [1,W]$. It encompasses the position of the end-effector in Cartesian coordinates $(x,y,z)$, the orientations $(r,p,y)$, and the gripper's joint angle $g$. For all the tasks, the input modalities consist of two modalities: ${\bf o} \in \mathbb{R}^{224 \times 224 \times 3}$, corresponds to a RGB image. ${\bf c}$ refers to a language embedding derived from natural language sequences.
Task Properties in \Cref{tab:data_summary} includes distractors (Dis), number of demonstrations (Dem), number of actions (Act), and whether high-precision is required (HiPrec).



\textbf{Duck Scooping:} Inspired by~\cite{antonova2023rethinking}, we explore the interaction between the policy and fluid dynamics.
In this task, we equip the robot with a ladle and the robot's objective is to scoop the duck out of the water. As depicted in the bottom right of \Cref{fig:task_teaser}(a), this task presents challenges due to perturbations caused by the entry of the ladle into the water. The flow of water affects the position of the rubber duck, necessitating the robot to execute precise and cautious movements in order to successfully capture the duck.

\textbf{Drum Beats}: This task is specifically designed for robots to learn periodic motions, a challenging feat due to the unique action representation required~\cite{yang2022learning}. As illustrated in \Cref{fig:task_teaser}(b), the task presents difficulty as the robot must accurately count the number of drum beats and determine when to cease drumming. A total of 150 expert demonstrations were obtained by teleoperating the robot to strike the drum three times in each demonstration.

\section{Evaluation} 
\label{sec:evaluation}
The efficacy of the proposed policy is evaluated through two experiments as described in \Cref{sec:tasks}. These experiments aim to address the following questions:
(a) Can the Diff-Control policy demonstrate generalization capabilities across diverse tasks?
(b) To what extent does the Diff-Control policy outperform the current state-of-the-art methods in terms of overall performance?
(c) What are the distinguishing characteristics and benefits of utilizing a stateful policy compared to a non-stateful policy?

We propose the 4 baselines: 1) Image-BC~\cite{jang2022bc}, 2) ModAttn~\cite{zhou2022modularity}, 3) BC-Z LSTM, and 4) Diffusion Policy~\cite{chi2023diffusion}. These baselines all adopt an image-to-action agent framework with varied implementation details.
For all experiments, we present the results obtained from the best performing configuration of each baseline method. All the baseline models are reproduced and trained using the collected expert demonstrations for a total of 3,000 epochs. Throughout the training process, checkpoints are saved every 300 epochs. In our analysis, we report the best results achieved from these saved checkpoints for each baseline method.

\begin{table}[ht]
\begin{center}
\caption{Results evaluation in forms of success rate (\%) and duration (sec) during policy execution}
\scalebox{0.99}{
\label{Tab:results2}
\begin{tabular}{c   c c c c }
    \toprule

     \multirow{2}{3em}{Method}
     &\multicolumn{2}{|c}{Duck Scoop}
     &\multicolumn{2}{|c}{Drum Beats}
     \\
     &\multicolumn{1}{|c}{Scoop}
     &Duration
      &\multicolumn{1}{|c}{Hit $\times$ 3}
     &\multicolumn{1}{c}{Duration}
     \\
    \midrule
     \multicolumn{1}{c|}{Image-BC}
     & 16\%
     & 21.04$\pm$1.99
     & \multicolumn{1}{|c}{0\%}
     & 19.50$\pm$0.79
     \\
     \multicolumn{1}{c|}{ModAttn}
     & 12\%
     & 20.43$\pm$0.16
    & \multicolumn{1}{|c}{0\%}
     &25.74$\pm$0.68
     \\
      \multicolumn{1}{c|}{BC-Z LSTM}
     & 36\%
     & 21.21$\pm$1.54
    & \multicolumn{1}{|c}{0\%}
     &17.53$\pm$0.31
     \\
    \multicolumn{1}{c|}{Diffusion Policy}
     & 76\%
     & 21.72$\pm$2.17
    & \multicolumn{1}{|c}{24\%}
     & 20.60$\pm$1.64
     \\
    \multicolumn{1}{c|}{\cellcolor[HTML]{E6F4FF} \textbf{Diff-Control}}

     & \cellcolor[HTML]{E6F4FF}\bf 84\%
     & \cellcolor[HTML]{E6F4FF}23.12$\pm$3.03
     & \multicolumn{1}{|c}{\cellcolor[HTML]{E6F4FF} \bf 72\%}
     &\cellcolor[HTML]{E6F4FF} 21.46$\pm$1.97
     \\
\bottomrule
\multicolumn{5}{l}{Means$\pm$standard errors} \\
\end{tabular}}
\end{center}
\vspace{-0.1in}
\end{table}

\begin{figure}[t]
\centering
\includegraphics[width=\linewidth]{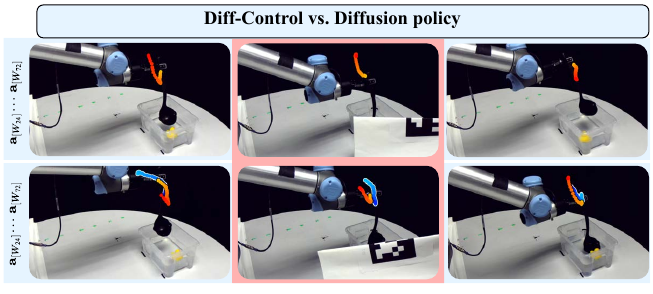}
\caption{When visual occlusion is applied, Diff-Control (2nd row) manages to scoop the duck from the water successfully. In contrast, the diffusion policy (1st row) fails. Red shade implies perturbation. }
\label{fig:tomato_recover}
\vspace{-0.15in}
\end{figure}

\subsection{Duck Scooping Evaluation}
\label{sec:duck_result}
In this task, we tested if Diff-Control policy is able to generate consistent actions in a dynamic setting. The success rate and task duration for the given task are presented in \Cref{Tab:results2}. Task duration was recorded as the time interval starting from the policy initiation until the duck was completely removed from the water. The experiment was conducted over 25 trials, with the duck randomly placed in the water for each trial. 
Among the state-of-the-art methods, the Diff-Control policy achieved a commendable success rate of 84\% while performing in this dynamic task. Notably, the Diff-Control policy demonstrated a tendency to successfully scoop the duck out in a single attempt, reaching a low enough position for accurate scooping. In contrast, the Image-BC and ModAttn failed often as the robot struggled to lower the ladle enough to reach the duck.
Diff-Control shows robustness against visual perturbation such as occlusion. In \Cref{fig:tomato_recover} (right), when the view is blocked, the diffusion policy fails immediately. However, Diff-Control can successfully scoop the duck without relying on any visual information because it learns an internal action transition to maintain a stateful behavior.

 \begin{figure} [ht]
    \centering
    \includegraphics[width=0.86\linewidth]{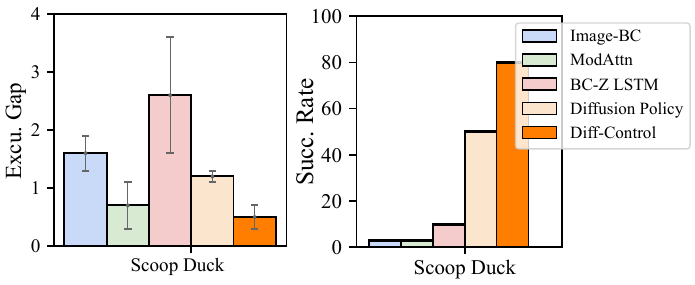}
    \caption{\textbf{Left}: Analysis of the execution gap (measured in cm) for duck scooping tasks during evaluation. \textbf{Right}: Assessment of the success rate for each policy when faced with perturbations in the duck scooping task.}
    \label{fig:exe_gap}
\vspace{-0.15in}
\end{figure}

\textbf{Robustness Evaluation}: Further evaluation was conducted to assess the consistency of the Diff-Control policy in this task. One way of quantifying action consistency is to measure the execution gap, which is the distance between the tail and head of two consecutive execution windows. In \Cref{fig:exe_gap}(left), the 3D distance of the execution gap is illustrated for each policy network, with Diff-Control displaying the smallest gap.
Furthermore, a supplementary set of experiments was carried out for the duck scooping task under visual occlusion. As depicted in \Cref{fig:exe_gap}, Image-BC and ModAttn only achieved a 0\% success rate when faced with perturbation, while BC-Z LSTM succeeded in 1 out of 10 trials in this scenario. Despite the diffusion policy achieving a 50\% success rate, Diff-Control demonstrated an 80\% success rate, showcasing the benefits of its stateful characteristic.
 \begin{figure} [t]
    \centering
    \includegraphics[width=0.95\linewidth]{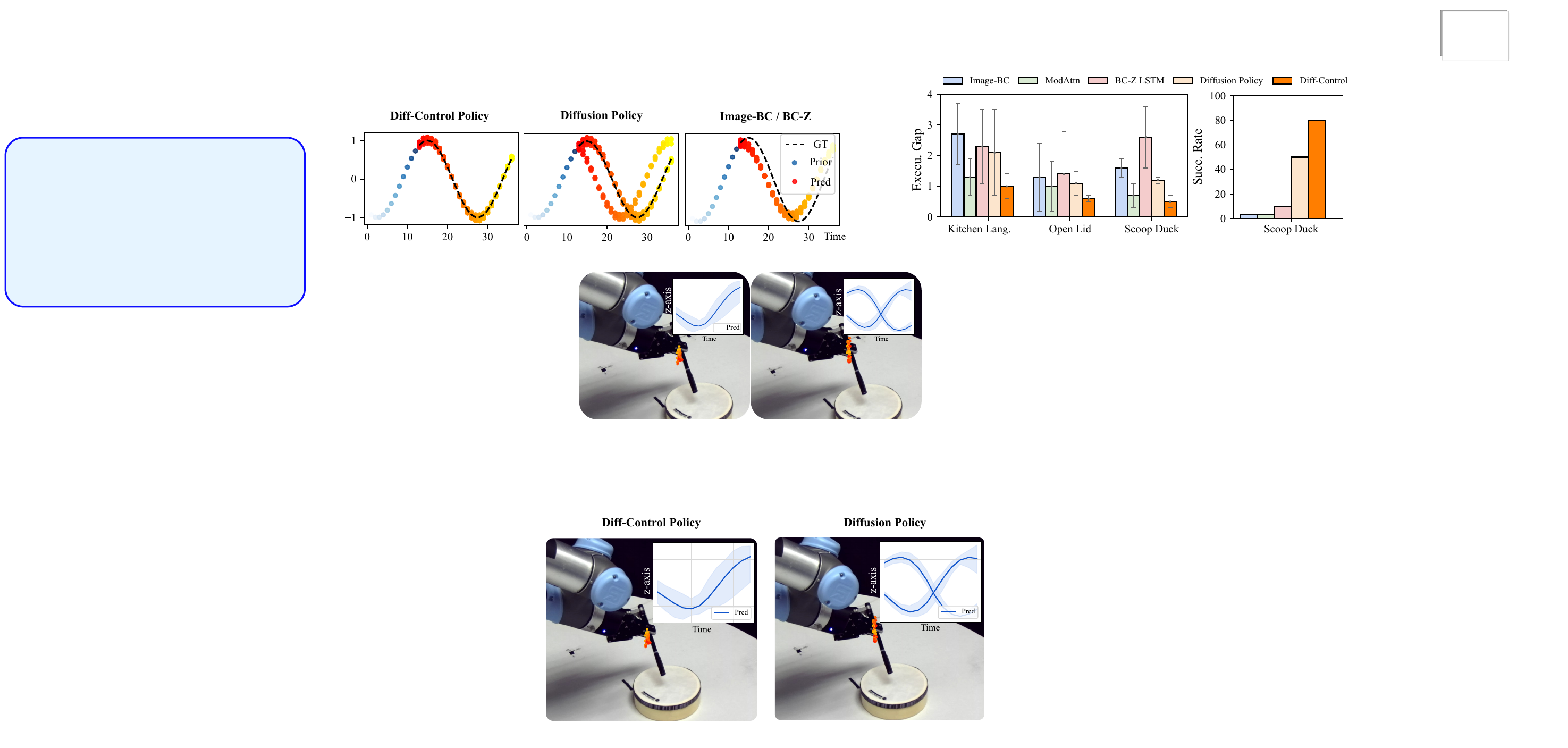}
    \caption{Generated action with variance from Diff-Control and Diffusion Policy during inference time.}
    \label{fig:drum_result}
\vspace{-0.15in}
\end{figure}
\subsection{Drum Beat Evaluation}
\label{sec:drum_result}

Each policy network was assessed in this experiment to evaluate the impact of statefulness versus non-statefulness on robots learning periodic motions. During testing, success was specifically defined as the robot hitting the drum \textbf{exactly three times} and then stopping. The results are presented in \Cref{Tab:results2}, with Diff-Control achieving the highest success rate of 72\%. This success rate surpasses the diffusion policy by 48\%. The majority of the baseline methods exhibit poor performance (Image-BC, ModAttn, BC-Z LSTM with 0\% success rate) due to their inability to accurately predict the direction of actions, such as the end-effector moving upward instead of downward, and the lack of appropriate halting actions. Consequently, The robot is unable to keep track of the number of times it strikes the drum and continues to strike the drum non-stop. While the BC-Z LSTM can accurately count the number of hits, it encounters difficulties in generating reliable actions initially. We visualize one test trial in the last row of \Cref{fig:task_plot}, where the last plot shows robot stopped after hitting the drum for 3 times.

Furthermore, we compared Diff-Control and the diffusion policy during the inference period, as shown in \Cref{fig:drum_result}. We sampled 10 action trajectories from each policy. Interestingly, the diffusion policy, without any prior actions, produced two distributions along the z-axis. This suggests that the policy struggled to determine whether it should descend to strike the drum or ascend after hitting the drum. In contrast, Diff-Control successfully generated actions by striking the drum. This observation shows using \textbf{Diff-Control as a stateful policy is beneficial for robot learning periodic behaviors}. 


\section{Conclusion} 
\label{sec:conclusion}
This study introduces Diff-Control, a stateful action diffusion policy designed for consistent action generation. The study explores the integration of diffusion model with ControlNet for robot action generation, demonstrating how temporal consistency can be enforced to enhance robustness and success rates. Furthermore, our findings underscore the robustness and effectiveness of Diff-Control in managing dynamic and stateful tasks while remaining resilient against perturbations.





\bibliographystyle{IEEEtran}
\scriptsize{
\bibliography{references}
}

\end{document}